\documentclass{article} 
\usepackage{colm2024_conference}
\usepackage{tabularx} 
\usepackage{microtype}
\usepackage{hyperref}
\usepackage{url}
\usepackage{booktabs}
\definecolor{darkblue}{rgb}{0, 0, 0.5}
\hypersetup{colorlinks=true, citecolor=darkblue, linkcolor=darkblue, urlcolor=darkblue}
\usepackage{listings}
\usepackage{graphicx}
\usepackage{xcolor}
\usepackage{times}
\usepackage{latexsym}
\usepackage{amssymb}
\usepackage{inconsolata}
\usepackage{xcolor}
\usepackage{fdsymbol}
\usepackage[T1]{fontenc}
\usepackage{times}
\usepackage{latexsym}
\usepackage{changepage}
\usepackage{multirow}
\usepackage{booktabs}
\usepackage{xspace}
\usepackage{amsfonts}
\usepackage{enumitem}
\usepackage{amsmath}
\usepackage{amssymb}
\usepackage{microtype}
\usepackage{pifont}
\usepackage{tikz}
\usepackage{colortbl}
\usepackage{url}
\usepackage{dsfont}
\newcolumntype{L}[1]{>{\raggedright\let\newline\\\arraybackslash\hspace{0pt}}m{#1}}

\usepackage[utf8]{inputenc}

\usepackage{microtype}
\definecolor{solve}{RGB}{104,78,167} 
\definecolor{explain}{RGB}{74,134,232} 
\definecolor{deepgreen}{RGB}{56,118,29}
\definecolor{deepred}{RGB}{164,43,38}
\lstdefinestyle{python}{
  language=Python,
  basicstyle=\ttfamily\small,
  keywordstyle=\color{blue},
  commentstyle=\color{gray},
  stringstyle=\color{red},
  showspaces=false,
  showstringspaces=false,
  tabsize=4,
  breaklines=true,
  aboveskip=-5pt,
  belowskip=-7pt
}
\usepackage{listings}
\definecolor{codegreen}{rgb}{0,0.6,0}
\definecolor{codegray}{rgb}{0.5,0.5,0.5}
\definecolor{codepurple}{rgb}{0.58,0,0.82}
\definecolor{backcolour}{rgb}{0.95,0.95,0.92}
 
\lstdefinestyle{mystyle}{
    language=Python,
    basicstyle=\ttfamily\tiny,
    backgroundcolor=\color{backcolour},   
    commentstyle=\color{codegreen},
    keywordstyle=\color{magenta},
    numberstyle=\tiny\color{codegray},
    stringstyle=\color{codepurple},
    breakatwhitespace=false,         
    breaklines=true,                 
    captionpos=b,                    
    keepspaces=true,                 
    numbers=none,                    
    numbersep=5pt,                  
    showspaces=false,                
    showstringspaces=false,
    showtabs=false,                  
    tabsize=2
}
 
\makeatletter
\newcommand\semismall{\@setfontsize\semismall{8}{8}}
\newcommand\dbsmall{\@setfontsize\semismall{7}{7}}
\usepackage{array}
\newcolumntype{C}[1]{>{\centering\arraybackslash}p{#1}}

\title{Distilling Algorithmic Reasoning from LLMs \\ via Explaining Solution Programs}

\author{Jierui Li  ~\& ~Raymond Mooney  \\
The University of Texas at Austin\\
\texttt{\{jierui, mooney\}@cs.utexas.edu} 
}

\colmfinalcopy 
\begin{document}

\maketitle

\begin{abstract}
Distilling explicit chain-of-thought reasoning paths has emerged as an effective method for improving the reasoning abilities of large language models (LLMs) across various tasks. However, when tackling complex tasks that pose significant challenges for state-of-the-art models, this technique often struggles to produce effective chains of thought that lead to correct answers. In this work, we propose a novel approach to distilling reasoning abilities from LLMs by leveraging their capacity to explain solutions. We apply our method to solving competitive-level programming challenges. More specifically, we employ an LLM to generate explanations for a set of \textit{<problem, solution-program>} pairs, then use \textit{<problem, explanation>} pairs to fine-tune a smaller language model, which we refer to as the Reasoner, to learn algorithmic reasoning that can generate "how-to-solve" hints for unseen problems. Our experiments demonstrate that learning from explanations enables the Reasoner to more effectively guide the program implementation by a Coder, resulting in higher solve rates than strong chain-of-thought baselines on competitive-level programming problems. It also outperforms models that learn directly from \textit{<problem, solution-program>} pairs. We curated an additional test set in the CodeContests format, which includes 246 more recent problems posted after the models' knowledge cutoff.
\end{abstract}

\section{Introduction}

Recent Large Language Models (LLMs) have shown impressive capabilities for various reasoning tasks, including multi-hop question answering \citep{wang-etal-2022-iteratively, lyu2023faithful}, symbolic reasoning \citep{hua-zhang-2022-system}, and math word problem-solving \citep{chen2022program,zhou2023leasttomost}. 
Chain-of-thought (CoT) prompting \citep{wei2022chain} addresses limitations of previous LLMs by instructing them to generate intermediate steps towards the final answer, thereby decomposing complex problems step-by-step.

However, challenges remain, particularly in complex reasoning tasks like algorithmic programming. For example,  the majority of human competitors still outperform advanced models like GPT-4 in Codeforces contests~\citep{openai2023gpt4}. Complex programming problems have stringent time and space complexity constraints, where straightforward implementation methods like \citep{chen2021evaluating, yin2018learning, hendrycks2021measuring}, often yield time-consuming brute-force solutions. 


A number of efforts have been made to tackle this challenging task \citep{Li_2022, zhang2023algo, olausson2023selfrepair, ridnik2024code} by adding extra clustering or verification steps to filter or iteratively refine generated programs. While those methods focus on flow engineering for code generation,  there have been limited attempts to explicitly enhance models' intrinsic reasoning abilities in this context.  

\begin{figure*}
	\centering
		\includegraphics[width=0.98\textwidth]{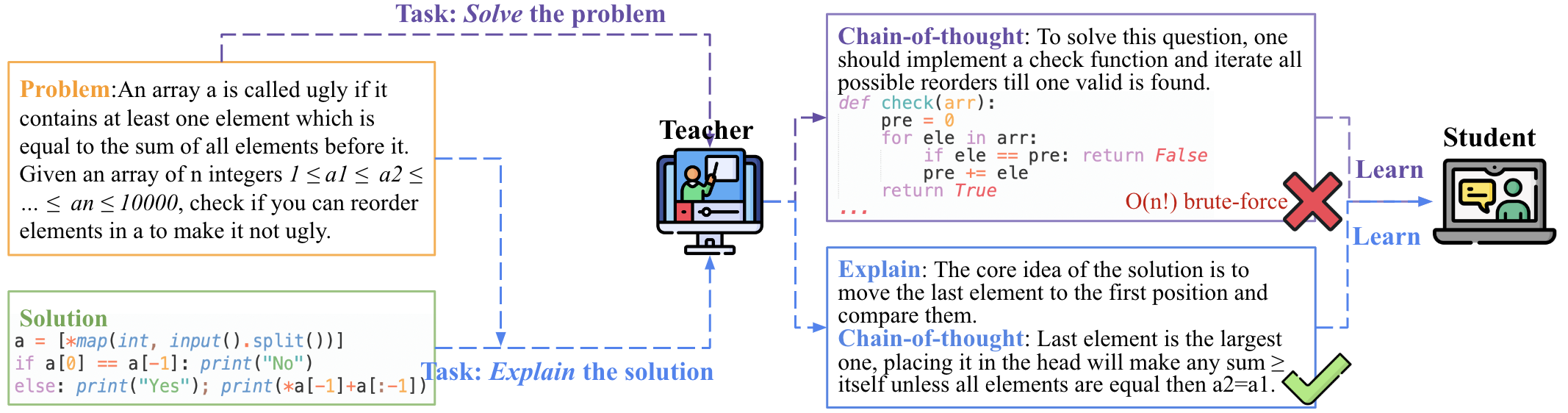}
	\caption{Comparison between \textcolor{solve}{Solve-based} and \textcolor{explain}{Explain-based} chain-of-thoughts distilling. Top:  \textcolor{solve}{Solve-based CoT distilling} is likely to generate incorrect or inefficient solutions. Bottom: \textcolor{explain}{Explain-based CoT distilling} can generate high-quality reasoning processes by explaining the oracle solution.}
	\label{fig:distill}
\end{figure*}

Human-written rationales for solving algorithmic reasoning problems, known as \textit{editorials}, are hard to collect as they are often posted on personal blogs or as tutorial videos. An alternative is to distill such natural-language-described problem-solving strategies from larger models. Distilling explicit chain-of-thoughts (CoT) reasoning processes has been shown as an effective method to learn multi-step reasoning from larger models \citep{hsieh2023distilling, yue2023mammoth}. Usually, a teacher model is required to solve a set of problems while giving CoT reasoning paths at the same time, as illustrated in Figure \ref{fig:distill}. However, when facing challenging tasks where state-of-the-art models struggle to generate effective solutions\footnote{Experiments show that only 12\% of GPT-4's generated solutions are correct given 200 problems randomly sampled from the CodeContests training set.}, it becomes infeasible to gather reasoning processes at scale.

To tackle this problem, we propose to utilize large language models' capacities in understanding and explaining solutions rather than solving problems. Specifically, we leverage a state-of-the-art LLM to read both the problem statement and an oracle human-written solution program, and generate an \textit{editorial}-style chain-of-thought on how to solve the problem by explaining the solution. Then, a student LLM learns the algorithmic reasoning processes from the explicit chain-of-thought paths. We compare our explain-based CoT distilling method with solve-based CoT distilling in Figure \ref{fig:distill}. While solve-based CoT distilling requires the teacher model to reach the correct and efficient solution to hard problems, our explain-based distilling only requires the teacher model to faithfully explain the correct solution. Since explaining competitive-level code is a feasible task for strong LLMs \citep{li2023explaining}, explain-based distilling can yield CoT reasoning processes with high quality and less noise. 

\begin{figure*}
	\centering
		\includegraphics[width=0.85\textwidth]{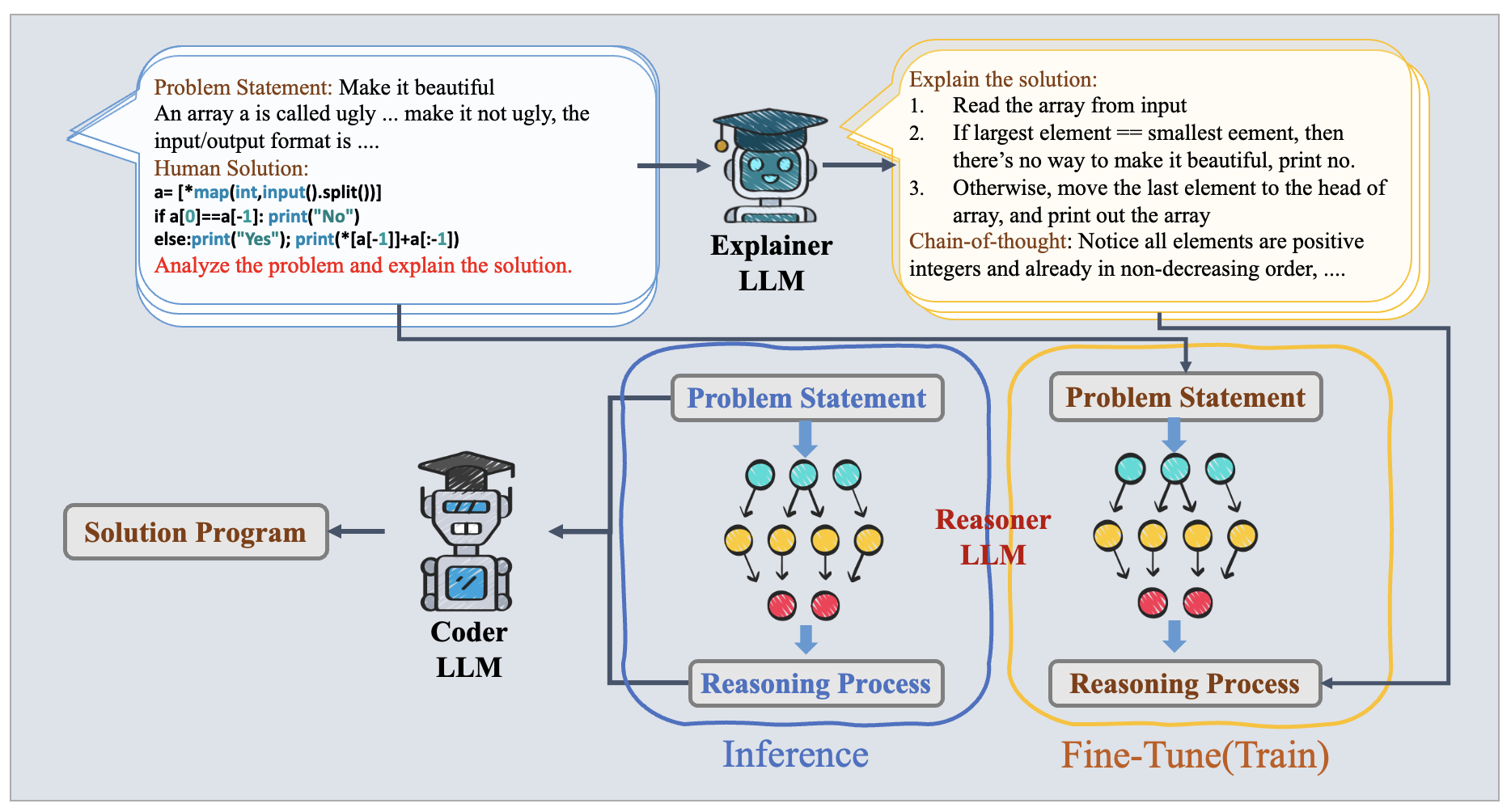}
	\caption{The framework of our approach. We use \textbf{Explainer} LLM to generate explanations given \textit{<problem, solution-program>} pairs; then train \textbf{Reasoner} LLM to generate explanations given problem statements. During inference time, given the problem, the \textbf{Reasoner} can generate a reasoning process in the same format as solution explanations, which could be provided to the \textbf{Coder} as a hint to solve the problem better.}
	\label{fig:main method}
\end{figure*}

We further proposed a reason-then-implement framework to solve algorithmic reasoning problems, utilizing the proposed explain-based CoT distilling. The framework consists of 3 major components: 1) an \textbf{Explainer} to annotate explanations for a set of \textit{<problem, solution-program}>; 2) a \textbf{Reasoner} to learn to generate intermediate reasoning processes for a given problem; and 3) a \textbf{Coder} to implement the solution for an unseen problem {\it given} the output from the \textbf{Reasoner}. The framework and its fine-tuning and inference stages are presented in Figure \ref{fig:main method}.

The resulting usage of this Reasoner fine-tuned with explanations provides superior performance over direct code generation techniques and strong zero-shot chain-of-thought prompting baselines. Experiments on open and closed models also show that compared to direct learning on public-available \textit{<problem, solution-code>} pairs, models learned on explicit natural language reasoning processes can generalize better to unseen problems.

Decomposing algorithmic problem-solving into separate \textbf{Coder} and \textbf{Reasoner} modules surpasses the effectiveness of models fine-tuned directly on code. The success of our approach is rooted in the nuanced semantic richn meaning of natural language. The proposed method relies solely on providing better problem reasoning to instruct the implementation, which can be combined with different \textbf{Coder} and pipelines such as self-debugging \citep{chen2023teaching}.

To summarize, our work makes the following contributions:
\vspace*{-2mm}
\begin{enumerate}
\setlength\itemsep{-1pt}
    \item We introduce an alternative, adaptable framework to distill complex reasoning processes from LLMs: instead of having LLMs solve problems, it leverages LLMs to explain solutions. 
    \item Our proposed distilling method is data-efficient: by fine-tuning with 8248 data points, 8M tokens in total, consistent performance gains in solve rates are observed on GPT-3.5 and Deepseek coder 7b. 
    \item For fair evaluation, we propose a new test set extracted from Codeforces. We advocate to take code efficiency into consideration in addition to code correctness. 
\vspace*{-1mm}

\end{enumerate}

\section{Related work}

\subsection{Solving algorithmic-reasoning programming problems}
Early endeavors in applying deep learning to tackle algorithmic reasoning in programming problems, such as those by \citet{balog2017deepcoder}, primarily utilized traditional approaches like SMT solvers and heuristic search. These methods were adept at generating short programs for relatively simple problems. Expanding on this, \citet{DBLP:journals/corr/abs-1802-04335} curated a dataset of human-written problem statements and solutions for Codeforces challenges. They introduced baseline sequence models capable of solving a modest portion of this dataset. The advent of Transformer-based models, such as Alphacode \citep{Li_2022}, dramatically improved performance on these challenges. Subsequent advancements by \citet{zelikman2023parsel} and \citet{olausson2023selfrepair}, as well as \citet{zhang2023algo}, further refined code generation techniques. They integrated additional layers for clustering or verification, enhancing the models' ability to filter and iteratively refine the generated programs, thereby improving both accuracy and efficiency.

\subsection{Distilling reasoning abilities}
Knowledge distillation, particularly from large models to smaller and more efficient counterparts, has been extensively studied \citep{Bucila2006ModelC, hinton2015distilling}. The advent of LLMs has further intensified research in this area, with a focus on distilling knowledge from LLMs into smaller models. This has often been coupled with the curation of LLM datasets \citep{taori2023alpaca, hsieh2023distilling}. Specifically, \citet{hsieh2023distilling} employed CoT prompting to extract rationales from LLMs for training smaller models in a multi-task learning setting. Similarly, \citet{yang2023effective} and \citet{zhu2023pad} distilled reasoning abilities from LLMs using CoT or program-of-thought processes. These approaches typically treat data synthesis as analogous to the target task, leveraging advanced models for higher-quality output generation. Our method differs as we replace the problem-solving CoT generation with an oracle-solution-explaining task. It transforms the nature of the task to facilitate the generation of high-quality outputs in a more manageable framework.

\section{Method}
Our method employs a two-step framework: In the learning stage, an \textbf{Explainer} generates a set of chain-of-thought reasoning processes by explaining solutions, and a \textbf{Reasoner} is fine-tuned on this data to generate a reasoning process given only a problem description. In the inference stage, the \textbf{Reasoner} generates the reasoning process for an unseen problem, which, along with the problem, is passed to a 0-shot \textbf{Coder} to implement. We include sample prompts in appendix \ref{sec:app}. 

\subsection{Problem Formulation}
We formalize our task with a problem set consisting of $n$ problems $P=\{p_1, p_2,\cdots,p_n\}$; each problem $p_i$ is a text sequence that clearly describes the following aspects: 
\begin{itemize}
\vspace*{-3mm}
\setlength\itemsep{-1pt}
    \item Problem statement: a natural language description of the problem, as shown in the \textcolor{orange}{upper left box} in Figure \ref{fig:distill}.
    \item Input/Output: input/output formats and constraints for the submitted program $s$.
    \item Example: An example of a correct input/output pair.
    \item Note (Optional): Explanation of the Example input/output.
\end{itemize}
\vspace*{-2mm}
Each $p_i$ corresponds to a set of oracle human solutions $S_i=\{s_i^1, s_i^2, \cdots, s_i^t\}$ where $t$ is the number of total solutions of $p_i$. We then select the shortest program written in Python from $S_i$ and form the \textit{<problem, solution-program>} pair $<p_i, s_i>$.

\subsection{Explainer: Extracting reasoning processes through explaining solution programs}

The \textbf{Explainer} is tasked with explaining a solution program. It serves the role of Teacher in \textcolor{solve}{Solve-based CoT distilling}, as described in Figure \ref{fig:distill}. Given a pair, $<p_i, s_i>$, it generates an explicit reasoning process $e_i$ in natural language. This is inspired by how human competitors learn problem-solving skills from past problems: they learn by reading editorials, step-by-step guidelines on approaching and solving the problems. While human-written editorials are hard to collect or annotate at scale, we ask an LLM to automatically generate them. We design an editorial-style chain-of-thought reasoning template and ask the \textbf{Explainer} to follow it to explain solutions. 
Specifically, an editorial for an algorithmic reasoning problem refers to a comprehensive explanation or walk-through on how to solve a problem, which includes problem analysis, strategy development, solution explanation, time/space complexity analysis, etc.

We leverage the capabilities of LLMs to explain solution code to generate automated explanations for problem-solution pairs $<p_i, s_i>$. The \textbf{Explainer} is prompted to create a detailed explanation, $d_i$, for each pair. Our focus is predominantly on the reasoning process, encompassing the following critical aspects that are often included in human-expert-written editorials:
\label{sec:aspects}
\begin{enumerate}
\vspace*{-3mm}
\setlength\itemsep{-1pt}
    \item Problem Restatement: Summary and analysis of the problem.
    \item Conceptual Evolution: How one approaches the problem and the development of a problem-solving strategy.
    \item Key to Solution: The key idea behind the solution.
    \item Solution Description: Brief, verbal description of the solution, focusing on the high-level algorithm.
    \item Step-by-Step Solution Explanation: A more detailed description, focusing on the steps in the implementation.
    \item Common Pitfalls: Some common mistakes one could make when approaching the problem, or edge/corner cases to be considered.
\end{enumerate}
Based on the length of  $p_i$ and $s_i$, as well as the difficulty ratings (ranging from 800 to 3600), we filter the training set from \citet{Li_2022} to curate a dataset $\{<p_1, s_1>, \cdots,<p_n, s_n>\}$. Utilizing GPT-4-0613 as the Explainer, we generate ``silver'' explanations for these pairs, resulting in a comprehensive problem-solution-explanation dataset {$<p_1, s_1, d_1>$, ..., $<p_n, s_n, d_n>$}. After further cleaning, we have 8248 triplets in total.

\subsection{Reasoner: Fine-tuned to generate reasoning processes for problems}

Given the inherent diversity and potential lack of readability of the code available for algorithmic reasoning problems, our approach focuses on fine-tuning the Reasoner on problem statements $p_i$ and reasoning processes $d_i$, which are <$p_0$, $d_0$>, <$p_1$, $d_1$> $\cdots, <p_n, d_n>$. These problems and reasoning processes, rich in semantics, encapsulate the essential steps for problem resolution, including the algorithms used and specific problem-solving approaches employed. The \textbf{Reasoner} serves the role of Student in \textcolor{solve}{Solve-based CoT distilling}, as illustrated in Figure \ref{fig:distill}.

We adopt a weighted fine-tuning strategy, with simpler, more recent problems weighted more heavily during training. Overly challenging problems might lead to low-quality noisy explanations that hurt the training of the \textbf{Reasoner}. \label{sec:reasoner}Recent solutions usually feature more in-date implementations (e.g., Python 3 rather than Python 2), enhancing the code's interpretability.
We fine-tune an LLM on 8,248 \{$p_i$, $d_i$\} pairs, and then use the fine-tuned model as the final Reasoner. At the inference time, it generates $\hat{d_j}$ for the $j$th problem statement $p_j$. The generated reasoning process $\hat{d}$ can be considered as a natural-language ``hint'' given to the \textbf{Coder} to help solve the problem.


\subsection{Coder: Reasoner-Hinted Code Implementer}
Finally, we have a zero-shot Coder to generate code utilizing hints from the \textbf{Reasoner}. Since the focus of this work is enhancing \textit{algorithmic reasoning} rather than \textit{code implementation} abilities, we do not further fine-tune the \textbf{Coder}. As described earlier, the reasoning process $\hat{d_j}$ for problem $p_j$ contains a problem restatement, conceptual evolution, key to the solution, solution description, step-by-step solution explanation, and common pitfalls, making it a detailed hint to assist the implementation of the solution. We concatenate $<p_j, \hat{d_j}>$ as the input to produce the input for the Coder, which it then uses to analyze the problem together with the reasoning hint, to generate programs that solve the problem.

\section{Experiments}
\subsection{Experimental Setup}

\paragraph{Model}We use GPT-4-0613 ~\citep{oai2023chatgpt,openai2023gpt4} as the \textbf{Explainer} since it's the strongest model that we have access to. We also choose the strongest models with fine-tuning access as the \textbf{Reasoner} and \textbf{Coder}. For a closed model, we choose GPT-3.5-turbo-1106 (henceforth GPT-3.5); for an open model, we choose Deepseek Coder 7B (henceforth Deepseek 7b) \citep{guo2024deepseekcoder}.\footnote{\url{https://huggingface.co/deepseek-ai/deepseek-coder-7b-instruct-v1.5}} The temperature $t$ is set to $0.5$ when sampling multiple(>1) is needed in our main experiments. The context window is set to 4096 and we 
truncate single examples with more than 4096 tokens.
\paragraph{Data}As suggested by \citet{huang2023competitionlevel}, to test the capacity of closed models like GPT, using recently-released data can reflect the model's performance more faithfully. To ensure novel test data given that GPT-3.5 was trained on data including problems and solutions from Codeforces, we extracted 246 recent problems from Codeforces as our main test set, \textit{CF Prob}. It contains 246 real online contest problems, the earliest of which dates to Aug.\ 2023, guaranteeing they were posted after the knowledge cutoff for GPT-3.5 (Jan.\ 2022). We also experimented with the 165 problems from the CodeContests benchmark \citep{Li_2022} test set, whose earliest problem was published in Oct.\ 2021. Table \ref{tab:dataset} gives statistics on the difficulty ratings of both sets. 

\begin{table}[h]
\vspace*{-2mm}
\centering
\scalebox{0.72}{
\begin{tabular}{l|c|c|c|c|c}\hline
& \multicolumn{5}{c}{Difficulty Ratings} \\ \hline
Dataset & total & \small{[800, 1000]} &\small{(1000, 1500]}& \small{(1500, 2000]} & \small{(2000, 3600]}\\ \hline
\small{CodeContests} & $165$ &$18.2\%$ & $17.0\%$ & $20.0\%$ & $44.8\%$\\ 
\small{CF Prob} & $246$ & $22.0\%$ & $23.6\%$ & $18.3\%$ & $36.2\%$\\ \hline
\end{tabular} }

\caption{Difficulty statistics (higher ratings = more difficult) for CodeContests and our proposed CF Prob. Both are sourced from Codeforces.}
\label{tab:dataset} 
\end{table}

\paragraph{Metric} We adapt pass@k \citep{chen2021evaluating} as our evaluation metric for solve rates. For each problem $p_i$, we sample $k$ programs generated by the \textbf{Coder} and evaluate them using \textbf{solve@k} which, when given to Codeforces' online judge, measures the percentage of problems solved by having at least \textit{one} program in the top $k$ that passes all hidden tests. 
We first filter in programs by executing and examining their output on the public test cases before submitting them. 
Measuring solve rates using the online judge provides a fairer comparison between models and human participants as it also requires efficiency in solutions, rejecting brute-force solutions when an efficient algorithm exists for a problem. 

We have models generate $10$ programs to compute solve@10, then compute solve@1 and solve@5 by calculating the probability of having at least one solution in the top $k$:
$$
\textit{solve@k} = \frac{1}{n}\sum_{i=1}^{n}P_i ~; ~~ 
P_i = 1 - \frac{\binom{10 - g_i}{k}}{\binom{10}{k}} \textit{ if } 10-g_i>k \textit{ else 0}
$$
where $g_i$ is the number of ``pass'' programs in 10 tries for $i$th problem. 
\subsection{Main Results}

\paragraph{Zero-shot Baselines}
\vspace*{-1.5mm}
\begin{itemize}
\vspace*{-3mm}
    \item \textbf{Direct Prompting}  Given the problem statement $p_i$, directly prompt the model to generate solution programs. 
    \item \textbf{Naive Chain-of-thought} Adapt the well-known ``Let\'s think step-by-step'' CoT prompting proposed in \citep{kojima2022large}. 
    \item \textbf{Editorial Chain-of-thought} Instruct the model to analyze the problem following an editorial chain-of-thought style giving aspects mentioned in Section \ref{sec:aspects}.
    \item \textbf{Zero-Reasoner\&Zero-Coder} Use a 0-shot Reasoner to generate natural language reasoning processes as hints for the 0-shot Coder.
\end{itemize}

\paragraph{Fine-tuning Methods}
\vspace*{-1.5mm}
\begin{itemize}
\vspace*{-3mm}
    \item \textbf{Fine-tuned Coder} Fine-tune the \textbf{Coder} to generate code given the problem, using the same 8248 \textit{<problem, solution-program>} pairs from CodeContests \citep{Li_2022}.
    \item \textbf{Fine-tuned Reasoner \&Zero-Coder } Use the Reasoner fine-tuned on solution explanations to generate natural-language reasoning hints for the 0-shot \textbf{Coder}. In addition to using the \textit{full} reasoning processes. We also separately tested the effect of each editorial aspect (e.g., common pitfalls) and selected the highest performing aspect as the \textit{best} aspect.
\end{itemize}

The results are presented in Table \ref{tab:main-eval}. With our fine-tuned \textbf{Reasoner}, the \textbf{Coder} achieves the best solve@10 and solve@5 rates across open/closed models and two datasets, supporting the effectiveness of our method. Compared to using the original model as the Reasoner (0-shot Reasoner), the best aspect (Step-by-Step) from the reasoning process alone can boost solve@10 from 3.3\% to 6.1\%, suggesting that our fine-tuning method does distill some algorithmic reasoning ability into the student model. 

Note that solely fine-tuning \textit{<problem, solution-program>}pairs actually hurts performance. One reason might be that solutions to Codeforces problems are often written under a time constraint with poor readability, making it difficult for the model to generalize given the limited amount of fine-tuning data. Since instruction-tuned models would generate intermediate thought processes by default, we experimented to explicitly forbid the model from any reasoning/analysis preceding implementation. Compared to the direct prompt, the solve@10 rate on CF Prob w/GPT-3.5 drops from 3.3\% to 2.0\%. Even with this setting, the generated code often contains comments and meaningful variable/function naming that are less in human competitors' code. This observation further supports our claim that meaningful, semantic-rich natural language can help the model generalize to unseen problems.

\textit{Unless unspecified otherwise, we use CF Prob and GPT-3.5 for experiments onwards.} \footnote{We use Deepseek Coder for tuning experiments and GPT-3.5 for all other experiments mainly considering their accessibility and efficiency.}

\begin{table}[ht!]
\centering
\scalebox{0.7}{
\begin{tabular}{lC{0.8cm}C{0.8cm}C{1.15cm}|C{0.8cm}C{0.8cm}C{1.15cm}|C{0.8cm}C{0.8cm}C{1.15cm}|C{0.8cm}C{0.8cm}C{1.15cm}}\hline
Dataset & \multicolumn{6}{c}{\textbf{CF-Prob}} & \multicolumn{6}{c}{\textbf{CodeContests}} \\
Coder/Reasoner Model& \multicolumn{3}{c}{GPT-3.5} & \multicolumn{3}{c}{Deepseek 7b} & \multicolumn{3}{c}{GPT-3.5} & \multicolumn{3}{c}{Deepseek 7b} \\ \hline
\multicolumn{13}{c}{Zero-shot Methods}\\ \hline
& \small\textbf{solve@1} &\small\textbf{solve@5}& \small\textbf{solve@10} & \small\textbf{solve@1} & \small\textbf{solve@5} & \small\textbf{solve@10} & \small\textbf{solve@1} &\small\textbf{solve@5}& \small\textbf{solve@10} & \small\textbf{solve@1} & \small\textbf{solve@5} & \small\textbf{solve@10}\\ \hline
Direct Prompt & $1.1$ &$2.7$ & $3.3$ & $0.4$& $0.9$& $1.6$& $1.9$ & $3.5$& $4.2$ & $0.8$& $1.2$ & $1.8$\\ 
Naive CoT & $1.2$ &$2.7$ & $3.3$ & $0.7$& $1.4$& $2.0$& $1.8$ & $3.7$& $4.8$ & $0.8$& $1.4$ & $1.8$\\ 
Editorial CoT & $1.1$ &$ 2.7$ & $3.6$ & $0.9$& $2.0$& $2.4$& $1.5$ & $3.7$& $4.8$ & $1.0$& $1.8$ & $2.4$\\ 
0-Reasoner Coder & $1.2$ &$2.5$ &$3.3$ & $0.7$& $1.9$& $2.4$ & $1.4$& $3.5$& $4.2$& $1.1$& $2.0$& $2.4$\\ \hline
\multicolumn{13}{c}{Fine-tuning Methods}\\ \hline
Fine-tune Coder & $0.5$ &$0.8$ & $1.6$ & $0.6$& $1.7$& $1.8$& $0.8$ & $1.2$& $1.8$ & $1.1$& $1.4$ & $1.9$\\  
\textbf{Fted Reasoner w/Full} & $1.1$ &$3.2$ & $4.9$ & $0.6$& $2.1$& $3.7$& $1.5$ & $4.2$& $6.1$ & $1.4$& $2.8$ & $3.6$\\ 
\textbf{Fted Reasoner w/Best} & $1.1$ &${3.7}$ &${6.1}$ & $1.0$& $2.9$& $4.1$& ${2.4}$ & ${5.6}$& ${7.2}$ & $1.4$& ${2.9}$ & ${4.2}$\\ 

~~~ \textit{Relative Increment} & & $+37\%$  & $ +69\%$  &
& $+53\%$ 
& $ +71\%$ 
&
& $ +51\%$ 
& $ +50\%$ 
&
& $+45\%$ 
& $ +75\%$ \\

\hline

\end{tabular}}
\caption{Performance of baselines and our methods. Fine-tuned Reasoner (Full) refers to using all aspects from the reasoning process, while Fine-tuned Reasoner (Best) only uses the best aspect. Relative Increment(by percentage) is what's over the best baseline. Solve@k: Solve rates(Percentage) when sampling k.} 
\label{tab:main-eval} 
\end{table}

\subsection{Effect of Tuning and Sampling} 
\paragraph{Fine-tuning Strategy}
As we mentioned in Sec.\ \ref{sec:reasoner}, harder problems are more challenging and frequently beyond the models' capacity. Therefore, the silver explanation given by the \textbf{Explainer} might not be as accurate, adding noise to the training of the Reasoner. So it might be more effective to focus fine-tuning on the simpler subset.
We compare three strategies to fine-tune deepseek-coder 6.7b: 1) Uniformly fine-tune on the entire dataset; 2) Fine-tune with simpler problems weighted more, and 3) Uniformly fine-tune on the simple subset only. The solve@10 rates as depicted in Table \ref{tab:sample}b, indicate that the strategy of just up-weighting simpler problems yields superior results. This suggests that in scenarios where the training data is noisy and limited, including noisier data while assigning greater weight to cleaner data can serve as a viable approach to enhancing generalization.

\paragraph{Program Sampling Strategy} Since our method partitions problem-solving between the Reasoner and Coder, it can adopt various sampling strategies. Specifically, the Coder can either a) implement different programs given one result from the \textbf{Reasoner} or b) implement one program for each of the different results from the \textbf{Reasoner}. Formally, to sample $k$ solutions, one can sample $k$ different results from the \textbf{Reasoner}, or alternatively, sample $k$ programs given only one reasoning result. For problem $p_i$ we sample $M$ reasoning results $\hat{d_i}^1, \hat{d_i}^2,\cdots, \hat{d_i}^M$ and $T$ programs $c_j$ for each process  $\hat{d_i}^j$, where $M\times T=k$. We show different choices of $M,T$ in Table \ref{tab:sample}. For cases where 1 deterministic sequence is needed (T=1 or M=1), we do not sample; otherwise, temperature is set to 0.5.

\begin{table}[h]
\centering
\scalebox{0.73}{

\begin{tabular}{l|c|c|c|c}\hline
Data Split& Data Size& Training Steps & Data Weight & solve@10\\ \hline
Simple+Hard & 8248  &1547 & Uniform & 3.3 \\ 
Simple+Hard & 8248  &1617 & 2xSimple:1xHard& \textbf{3.7} \\ 
Simple & 4691  & 1172 & Uniform & 3.3\\ 
\hline
\multicolumn{5}{c}{(a)fine-tuning strategy comparison w/Deepseek Coder} \\
\end{tabular}
\hfill ~ \hfill ~ \hfill
\begin{tabular}{l|c|c}\hline
Sample Reasoner & Sample Coder &solve@10\\ \hline
M=1 & T=10  &2.8 \\ 
M=2 & T=5  & 3.7 \\ 
M=5 & T=2  & 4.1\\ 
M=10 & T=1  & $\textbf{4.9}$\\ 
\hline
\multicolumn{3}{c}{(b)sampling strategy comparison w/GPT-3.5} \\
\end{tabular}
}
\caption{ (a)Fine-tune settings testing whether to up-weight simpler problems.  Training steps are kept similar unless data points are significantly fewer. (b)Sampling more from \textbf{Reasoner} vs. sampling more from \textbf{Coder}. Both solve@10(percent) are Fine-tuned Reasoner w/Full on CF Prob.}
\label{tab:sample} 
\end{table}

We found that for fixed $k = 10 = M \times T$, sampling more reasoning processes $M$ is more important than sampling more implementations for the same reasoning process $T$. The diversity in reasoning often reflects usage of different strategies or algorithms. Thus, sampling from the \textbf{Reasoner} potentially allows the \textbf{Coder} to try different methods for a problem. 
We therefore used $temp=0.5$ and $M=10, T=1$ for the experiments in Table \ref{tab:main-eval}(a) and the following section.

\subsection{Ablation Study} Since providing the full reasoning process with all aspects can distract the model, we found that providing a single aspect can yield a higher overall solve rate. We therefore studied the effects of using different aspects of the explanations generated by the \textbf{Reasoner}.
\begin{table}[h]
\centering
\scalebox{0.8}{
\begin{tabular}{l|c|c|c}\hline
& {Solve@1} &{Solve@5}& Solve@10 \\ \hline
*0-shot Reasoner & $1.2$ &$2.5$ &$3.3$ \\
w/ Full Reasoning process& $1.1$ &$3.2$ & $4.9$ \\ 
w/ Conceptual Evolution & $\textbf{1.5}$ &$\textbf{4.1}$ & $5.3$  \\
w/ Key to Solution &$1.0$ & $3.2$ & $4.5$\\
w/ Solution Description & $1.1$ & $3.5$ &$4.9$\\
w/ Common Pitfalls & $1.2$ & $3.4$ &$4.9$ \\
w/ Step-by-Step Solution Explanation & $1.1$ &$3.7$ &$\textbf{6.1}$ \\\hline
\end{tabular}}

\caption{Ablation Study on the different aspects described in Section \ref{sec:aspects}. Rather than fine-tuning another model, we give only each individual aspect of the \textbf{Reasoner}'s output to the {\bf Coder}. Solve@k: Solve rates(percentage) when sampling k. }
\label{tab:ablation} 
\end{table}

The results in Table \ref{tab:ablation} show that each aspect alone can achieve better solve rates against the un-fine-tuned \textbf{0-shot Reasoner}, proving the effectiveness of our explanation fine-tuning. Among all the different aspects, Conceptual Evolution achieves the best solve@1 and solve@5 while Step-by-Step Solution Explanation achieves the best solve@10.

Since Conceptual Evolution provides a solving strategy, it indicates how one should think about the problem and where to start, while also touching on the high-level idea behind the solution, as well as the choice of algorithm. Therefore, by giving a good starting point, it allows the \textbf{Coder} to follow up the idea and implement a correct solution. 
On the other hand, \textit{Step-by-Step Solution} focuses more on implementation details, giving specific instructions on what and how to implement, making it easy for the \textbf{Coder} to follow. However, it is not as robust as higher-level aspects: If the \textbf{Reasoner} makes a mistake in the details, the \textbf{Coder} may not spot the bug and produce an incorrect implementation.

\subsection{Program Submission Status} In our experiments, we find that, compared to our method, programs sampled with zero-shot methods pass more public tests, but yield significantly lower solve rates.  This indicates that the pre-trained large language models like GPT-3.5 and Deepseek Coder tend to give brute-force solution programs initially. We further study this phenomenon by analyzing the statistics of the statuses of submitted programs.

\begin{figure}
    \vspace*{-4mm}
    \centering
    \begin{minipage}[b]{0.58\textwidth}
    \includegraphics[width=\textwidth]{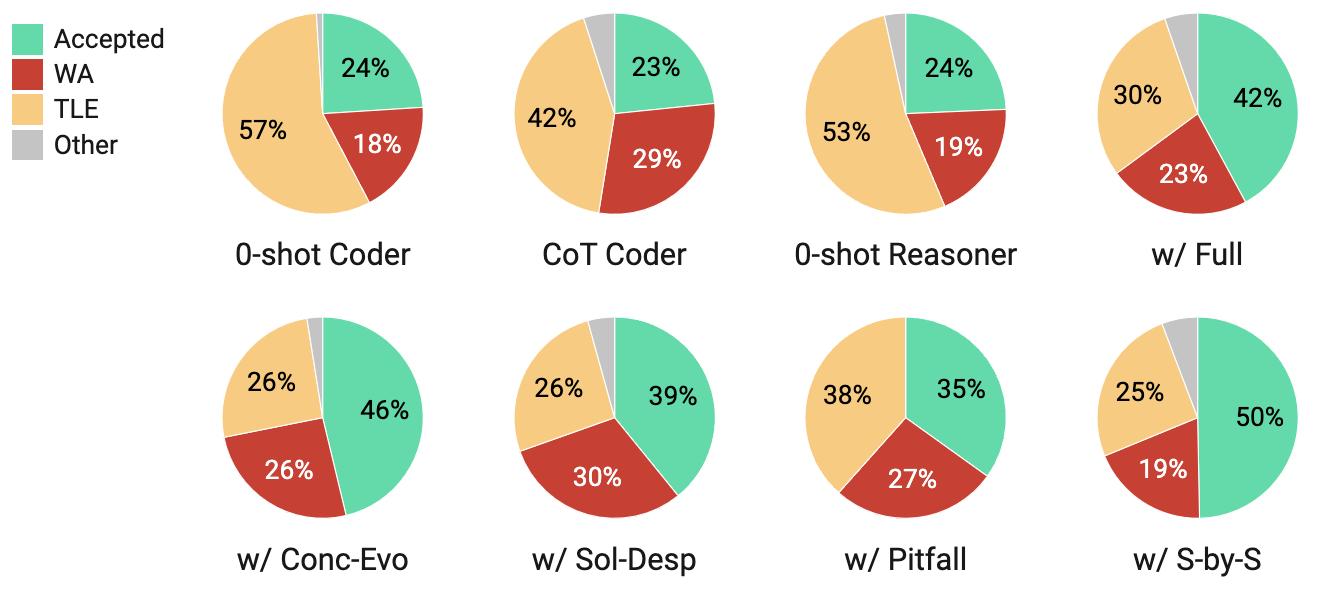}
    \vspace*{-4mm}
    \caption{Final online judgment of programs that pass public tests. Accepted: correct; TLE: time limit exceeded; WA: wrong answer(s) on private tests; Other: memory limit exceeded, runtime error etc.\protect\footnotemark}
    \label{fig:tle}
    \end{minipage}
    \hfill 
    \begin{minipage}[b]{0.38\textwidth}
    \includegraphics[width=\textwidth]{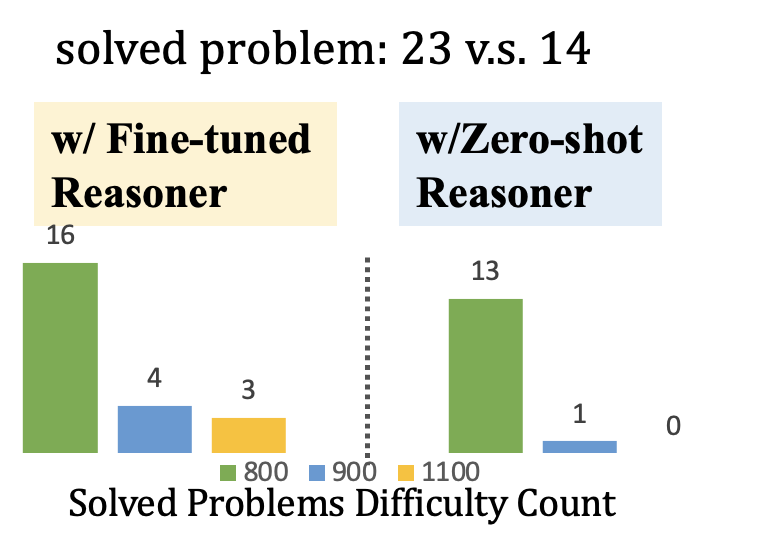}
    \vspace*{-4mm}
    \caption{The problem difficulty statistics for problems solved with fine-tuned or zero-shot \textbf{Reasoner} when sampling 100 reasoning processes per problem. }
    \label{fig:solvable}
    \end{minipage}
\end{figure}

In competitive-level programming, online judge systems give TLE (Time Limit Exceeded) to submissions that do not run efficiently enough to be considered correct. As illustrated in Figure \ref{fig:tle}, the baseline has $42\%$ to $57\%$ of its programs rejected due to time inefficiency. 
Our fine-tuned \textbf{Reasoner}, on the other hand, has learned to avoid brute-force strategies. With Step-by-Step Solution Explanation, $50\%$ of the programs that pass the public tests are accepted solutions, indicating that the Reasoner learns to use efficient algorithms rather than implementing brute-force solutions.

\footnotetext{This is for all submissions, i.e., one problem might have up to $k$ submissions, which is different from the problem-wise solve rate. }

\subsection{Solved Problems Analysis}
Given that compared to zero-shot settings, using the Fine-tuned Reasoner only helps solve several more problems, we investigate further whether this improvement is due to learning from explanations or is merely coincidental. Thus, we aggregated all solved problems from 5 experiments with different sampling strategies and temperatures with both a zero-shot baseline Reasoner and our fine-tuned Reasoner. In other words, for each problem, we evaluated 50 programs generated under the zero-shot setting and another 50 under the explanation Fine-tuned setting. marking problems solved by either to obtain a final set of 23 ``solvable'' problems.

Then, we measured solve@100 for this ``solvable'' subset. For each problem in the solvable set, we sample 100 explanations from both the Zero-shot and Fine-tuned Reasoners. By asking the same Coder to implement them, we found that the Explanation-Fine-tuned Reasoner achieves a \textbf{100\%} solve@100 while Zero-shot Reasoner only achieves \textbf{60.9\%} on this set.

The difficulty distribution of these problems is shown in Figure \ref{fig:solvable}. We observed that with a fine-tuned \textbf{Reasoner}, more difficult problems can be solved, while the zero-shot original model mainly solves problems with the lowest difficulty rating (800). This highlights that learning from explanations can enhance the \textbf{Reasoner}'s ability to address more complex problems.

\subsection{Case Study}

\begin{table*}[ht]
  \centering
    \begin{tabular}{|p{13.3cm}|}
    \hline
   \small \textbf{Problem Statement: } \small Yarik recently found an array $a$ of $n$ elements and became very interested in finding the maximum sum of a non-empty subarray. The subarray he chooses must have alternating parities for adjacent elements.
For example, $[1, 2, 3]$ is acceptable, but $[1, 2, 4]$ is not, as 2 and 4 are both even and adjacent.
You need to help Yarik by finding the maximum sum of such a subarray.
 \\ \hline
\small \textbf{Fine-tuned Reasoner's Reasoning Process\textcolor{deepgreen}{~~$\checkmark$}: }\\
    \small ... checking all possible subarrays would be inefficient. Recognizing that the problem exhibits \textcolor{deepgreen}{optimal substructure and overlapping subproblems}, \textcolor{deepgreen}{dynamic programming} becomes a natural choice. ... the maximum sum of a subarray ending at index $i$ can be \textcolor{deepgreen}{extended} by including $a[i]$ if it \textcolor{deepgreen}{maintains the alternating parity condition}. This insight leads to the iterative update of $dp$ and the tracking of the maximum sum. \\\hline
\small \textbf{Zero-shot Reasoner's Reasoning Process\textcolor{deepred}{~~$\times$}: }\\
\small To solve this problem, we can use a \textcolor{deepgreen}{dynamic programming} approach. We can iterate through the array and \textcolor{deepgreen}{keep track of the maximum alternating sum ending at each element}. We will \textcolor{deepred}{maintain two values}, one for the maximum alternating sum ending at the current element with an \textcolor{deepred}{odd index} and another for the maximum alternating sum ending at the current element with an \textcolor{deepred}{even index}.
\\
\hline

  \end{tabular}
  \vspace*{-2mm}
  \caption{An algorithmic reasoning problem from \textit{CF Prob} with reasoning processes from the fine-tuned and 0-shot \textbf{Reasoner}.}
  \label{tab:casestudy}
\end{table*}

Table \ref{tab:casestudy} shows a problem\footnote{\url{https://codeforces.com/contest/1899/problem/C} }that can be solved by 14\% of sampled programs from experiments with Fine-tuned \textbf{Reasoner} but none by the zero-shot Reasoner. We compare the output of our \textbf{Fine-tuned Reasoner} to that of the 0-shot \textbf{Reasoner}. The generated program and further analysis in sampled reasoning processes are given in Appendix \ref{sec:app}.

This problem can be solved using dynamic programming while specifying the rule of having ``alternating parities for adjacent elements''. The Fine-tuned \textbf{Reasoner} analyzes the inner logic of how to update the DP status and final answer correctly considering the condition.  By contrast,  the 0-Shot \textbf{Reasoner} only gives plausible surface-level reasoning output by connecting ``alternating parities'' to keywords like ``odd'' or ``even''.

\section{Conclusions and Future Work}
In this study, we propose to distill large language models' ability to explain solutions into reasoning abilities used to help solve problems. More specifically, we use an LLM (Teacher) to explain, in natural language, the reasoning processes for \textit{<problem, solution-program>} pairs and fine-tune a smaller LLM (Student) on such generated reasoning processes. Using a two-phase framework of reason-then-implement where the student model plays the role of a \textbf{Reasoner} to instruct a zero-shot \textbf{Coder} to generate the implementation. Experiments on real problems from Codeforces demonstrate that our explain-based distilling outperforms several strong 0-shot baselines as well as fine-tuning with code solutions alone. Even when absolute solve rates are low given how challenging the competitive-level algorithmic problems are, our method can significantly improve the performance and solve relatively 37\% to 75\% more problems than the strongest baselines across all experiments. Our training set for fine-tuning only contains 8248 data points and 8M tokens, and it suggests the efficiency of the proposed distilling method. Quantitative and qualitative analyses reveal that the fine-tuned Reasoner learns to avoid brute-force implementations and favors more efficient programs compared to its non-fine-tuned counterparts. We also curated an up-to-date test set for algorithmic reasoning challenges.

Automatically generating and learning from explanations of problem solutions to improve reasoning presents a promising path forward for tackling a broader array of complex reasoning tasks. Future investigations could extend this insight into domains where current LLMs struggle, such as abstract problem-solving in mathematics or logical deduction in law.

\section*{Ethics Statement}
This work focuses on proposing a method to enhance the reasoning ability of large language models on algorithmic reasoning. We will release the dataset \textit{CF Prob} and the explanations generated for training if it's not against OpenAI's policy. We recognize the potential ways our framework can be misused, particularly in the context of online coding competitions. We firmly oppose any application of our work that would facilitate dishonest practices including cheating.

Another issue is that the data is in English and Python language only, which might need translation if used with other languages. Given the current performance of our models and the nascent state of algorithmic reasoning automation, we assess that the immediate industry application of our findings is unlikely. Consequently, we foresee minimal risk of adverse effects stemming from premature deployment in real-world scenarios.

\bibliography{colm2024_conference}
\bibliographystyle{colm2024_conference}

\appendix
\newpage
\section{Appendix}
\label{sec:app}
\subsection{Generated Reasoning Process Analysis}
\begin{table}[ht]
  \centering
    \begin{tabular}{|p{7cm}|p{7.5cm}|}
    \hline
   \multicolumn{2}{|L{15cm}|}{\small \textbf{Problem Statement: } \small Yarik recently found an array $a$ of $n$ elements and became very interested in finding the maximum sum of a non-empty subarray. However, Yarik doesn't like consecutive integers with the same parity, so the subarray he chooses must have alternating parities for adjacent elements. For example, $[1, 2, 3]$ is acceptable, but $[1, 2, 4]$ is not, as 2 and 4 are both even and adjacent. You need to help Yarik by finding the maximum sum of such a subarray. }
\\ \hline

\small \textbf{Fine-tuned Reasoner Generated Process $\checkmark$: }\\
    \small ... checking all possible subarrays would be inefficient. Recognizing that the problem exhibits \textit{optimal substructure and overlapping subproblems}, \textit{dynamic programming} becomes a natural choice. ... the maximum sum of a subarray ending at index $i$ can be \textit{extended} by including $a[i]$ if it \textit{maintains the alternating parity condition}. This insight leads to the iterative update of $dp$ and the tracking of the maximum sum. &
\vspace*{-4mm}    
\small \textbf{Non Fine-tuned Reasoner Generated Process $\times$: }~~
\small To solve this problem, we can use a dynamic programming approach. We can iterate through the array and keep track of the maximum alternating sum ending at each element. We will maintain two values, one for the maximum alternating sum ending at the current element with an odd index and another for the maximum alternating sum ending at the current element with an even index.\\\hline
    \small \textbf{GPT-3.5 Coder Generated Program: }\\
\vspace*{-4mm}

\begin{lstlisting}[language=Python]
def max_alternating_subarray_sum(arr):
    n = len(arr)
    dp = [0] * n
    dp[0] = arr[0]
    for i in range(1, n):
        if arr[i] % 2 != arr[i-1] % 2:
            dp[i] = max(dp[i-1] + arr[i], arr[i])
        else:
            dp[i] = arr[i]
    return max(dp)

t = int(input())
for _ in range(t):
    n = int(input())
    arr = list(map(int, input().split()))
    result = max_alternating_subarray_sum(arr)
    print(result)
\end{lstlisting}

&
\vspace*{-4mm}

\begin{lstlisting}[language=Python]
def max_alternating_sum(arr):
    n = len(arr)
    dp_odd, dp_even = [0] * n, [0] * n
    dp_odd[0] = arr[0]
    dp_even[0] = 0
    for i in range(1, n):
        dp_odd[i] = max(dp_even[i-1] + arr[i], arr[i])
        dp_even[i] = max(dp_odd[i-1] - arr[i], 0)
    
return max(dp_odd[-1], dp_even[-1])
t = int(input())
for _ in range(t):
    n = int(input())
    arr = list(map(int, input().split()))
    result = max_alternating_sum(arr)
    print(result)
\end{lstlisting}
    \vspace*{-5mm}\\ \hline
  \end{tabular}
  \caption{An algorithmic reasoning problem from \textit{CF Prob} with reasoning processes from the finetuned and 0-shot \textbf{Reasoner}. The incorrect process leads to a failed solution(right side). }
  \label{tab:a1}
\end{table}
We ask a human expert to evaluate the generated reasoning processes by Fine-tuned and Zero-shot Reasoner and compare them. Unlike \ref{tab:casestudy}, we look at all 10 reasoning processes for both. We found that while the 0-shot model shows the correct general understanding that it can be solved as a dynamic programming problem, it fails to further reason meaningfully. 9 out of 10 reasoning processes suggest maintaining `MaxOdd` and `MaxEven` variables to consider odd and even positions separately. Only one reasoning process suggests a variant of Kadane's algorithm, but fails to understand what does `` must have alternating parities for adjacent elements'' means.

On the contrary, with the fine-tuned \textbf{Reasoner}, 3 out 10 samples realize `odd` and `even` should be considered as the condition to update the sum, and 2 out 10 use a greedy method instead. This observation indicates a more human-mimicking property: by learning from explanation, the \textbf{Reasoner} tends to try different strategies.

As shown in \ref{tab:a1}, the main difference between 2 programs is their used algorithm. The correct reasoning process can guide the \textbf{Coder} to implement the correct solution, suggesting a better understanding of the problem condition.

\subsection{Prompts and Examples}
\begin{figure*}
\vspace{-1.2cm}
\footnotesize
\noindent\rule{\textwidth}{1pt}
\textbf{System Prompt:} You are an expert in algorithm and programming.\\
\noindent\rule{\textwidth}{0.5pt}
\textbf{User Prompt:} Analyze the following competitive programming problem and its provided solution. Write an editorial to help students understand the problem-solving approach and strategy. Consider the problem's details, the solution's approach, and the idea behind it.\\
-----------\\
Problem:
In this problem, we only consider strings consisting of lowercase English letters. Strings $s$ and $t$ are said to be isomorphic when the following conditions are satisfied:\\
* $|s| = |t|$ holds.\\
* For every pair i, j, one of the following holds:\\
* $s_i = s_j$ and $t_i = t_j$.\\
* $s_i \neq s_j$ and $t_i \neq t_j$.\\
For example, `abcac` and `zyxzx` are isomorphic, while `abcac` and `ppppp` are not.
A string s is said to be in normal form when the following condition is satisfied: 
* For every string t that is isomorphic to $s, s \leq t $holds. Here $\leq$ denotes lexicographic comparison.

For example, `abcac` is in normal form, but `zyxzx` is not since it is isomorphic to `abcac`, which is lexicographically smaller than `zyxzx`.\\
You are given an integer N. Print all strings of length N that are in normal form, in lexicographically ascending order.\\
Constraints\\
* $1 \leq N \leq 10$\\
* All values in input are integers.\\
Input\\
Input is given from Standard Input in the following format:\\
N

Output\\
Assume that there are K strings of length N that are in normal form: $w_1, \ldots, w_K$ in lexicographical order. Output should be in the following format:

$w_1$
:
$w_K$

Examples\\
Input\\
1

Output\\
a

Input\\
2

Output\\
aa\\
ab\\
-----------\\
Below is an accepted solution. Analyze it in the context of the problem.
\vspace{-0.2cm}    
\begin{lstlisting}[language=Python]
n = list(map(int, input().split(' ')))[0]

def dfs(i, mx, n, res, cur = []):
	if i==n:
		res.append(''.join(cur[::]))
		return

	for v in range(0, mx + 1):
		dfs(i+1, mx+(1 if v==mx else 0), n , res, cur+[chr(v+ord('a'))])
res =[]
dfs(0,0,n,res)
for w in res: print(w)
\end{lstlisting}
\vspace{-0.2cm} 
Answer using the following format, with clear, detailed explanations. Use precise terms and avoid ambiguity. Use specific descriptions instead of vague expressions like "rearrange it" or "apply some operation". Specific instructions for each point are as follows in brackets"{}":\\
1). Problem Restatement: {Understand every aspect of the problem first. Summarize the problem statement to remove the narrative/storytelling or thematic elements like characters or background story, abstract it into a formal statement while describing constraints, input-output specifications.}\\
2). Step-by-Step Solution Explanation: {Explain the solution code step-by-step in an algorithm level instead of explaining code line-by-line. Focus on the algorithm rather than implementation details like how to read input.}\\
3). Solution Description: {Based on the understanding of both the problem and the solution, describe the solution approach verbally. Explain the solution and the high-level reasoning behind it. Explain the WHY of the core algorithms/data structures/problem modeling.}\\
4). Conceptual Evolution: {Given all discussed above, describe how one would arrive at this solution. This can include how to analyze and approach the problem, the choices of type of algorithm used (e.g., dynamic programming, greedy, graph theory), the intuition behind the approach, and why this approach works for this problem. This is a narrative of the problem-solving journey.}\\
5). Common Pitfalls: {Pitfalls in the problem description OR common errors that students might make while attempting the problem OR corner/edge cases or offset handled in the solution.}\\
6). Key to Solution:{Use one sentence to illustrate the "aha!" steps (key idea or trick) in the solution. Be concise, specific and informative.}\\
\vspace{-0.2cm}     
\noindent\rule{\textwidth}{1pt}
\caption{\textbf{Explainer Prompt Example}: Example of the prompt we are using to generate Explanations from \textbf{Explainer} (GPT-4). The problem and solution is from \citet{Li_2022}.}
    \label{fig:example_explainer}
\end{figure*}

\begin{figure*}
\vspace{-1.2cm}
\footnotesize
\noindent\rule{\textwidth}{1pt}
\textbf{System Prompt:} You are an expert in algorithm and programming.\\
\noindent\rule{\textwidth}{0.5pt}
\textbf{User Prompt:} Given the algorithmic reasoning problem below, analyze it first, then develop a verbal description of the solution. Focus on the idea rather than the implementation details.\\
Problem:
In this problem, we only consider strings consisting of lowercase English letters. Strings $s$ and $t$ are said to be isomorphic when the following conditions are satisfied:\\
* $|s| = |t|$ holds.\\
* For every pair i, j, one of the following holds:\\
* $s_i = s_j$ and $t_i = t_j$.\\
* $s_i \neq s_j$ and $t_i \neq t_j$.\\
For example, `abcac` and `zyxzx` are isomorphic, while `abcac` and `ppppp` are not.
A string s is said to be in normal form when the following condition is satisfied: 
* For every string t that is isomorphic to $s, s \leq t $holds. Here $\leq$ denotes lexicographic comparison.

For example, `abcac` is in normal form, but `zyxzx` is not since it is isomorphic to `abcac`, which is lexicographically smaller than `zyxzx`.\\
You are given an integer N. Print all strings of length N that are in normal form, in lexicographically ascending order.\\
Constraints\\
* $1 \leq N \leq 10$\\
* All values in input are integers.\\
Input\\
Input is given from Standard Input in the following format:\\
N

Output\\
Assume that there are K strings of length N that are in normal form: $w_1, \ldots, w_K$ in lexicographical order. Output should be in the following format:

$w_1$
:
$w_K$

Examples\\
Input\\
1

Output\\
a

Input\\
2

Output\\
aa\\
ab\\
\noindent\rule{\textwidth}{0.5pt}
\textbf{Assistant Prompt}
Let's walk through the problem and think about how to solve it first.\\Conceptual Evolution:\\To arrive at this solution, one must understand the definition of isomorphic strings and normal form. The key insight is that a string in normal form will use the earliest possible character ('a') and only introduce a new character when it cannot use an existing one. This insight leads to a recursive generation of strings, where at each step, we either reuse an existing character or introduce the next character in the alphabet. The DFS algorithm is a natural fit for this recursive construction process.\\Key to Solution:\\The \"aha!\" step is realizing that a string in normal form will always start with 'a' and introduce new characters sequentially only when necessary, which can be efficiently generated using a depth-first search.\\Solution Description:\\The solution uses a depth-first search (DFS) approach to generate all possible strings in normal form. It builds strings character by character, ensuring that each new character either matches a previous character or is the next unused character in the alphabet. This guarantees that the generated strings are in normal form and lexicographically minimal.\\Common Pitfalls:\\- Not understanding the definition of isomorphic strings and normal form could lead to incorrect string generation.\\- Failing to ensure that the strings are generated in lexicographic order might result in an incorrect output sequence.\\- Overlooking the base case in the recursive function could lead to infinite recursion or missing strings.\\- Not considering the constraint that N is at most 10 could lead to inefficient solutions that do not scale.\\Step-by-Step Solution Explanation:\\- Read the integer N from the input.\\- Define a recursive function `dfs` that will build the strings in normal form. This function takes the current position `i`, the maximum character index `mx` used so far, the string length `n`, a list `res` to store the results, and the current string `cur`.\\- If `i` equals `n`, it means we have constructed a string of length `n`. Add the current string `cur` to the results list `res` and return.\\- Iterate over all possible next characters, represented by their index `v` from 0 to `mx + 1`. The character to be added is determined by converting `v` to a character starting from 'a'.\\- For each character, recursively call `dfs` with the next position `i + 1`, updating `mx` to `mx + 1` if `v` equals `mx` (indicating that we are using a new character for the first time), and appending the new character to `cur`.\\- Initialize an empty list `res` to store the resulting strings.\\- Call `dfs` starting from position 0, with `mx` set to 0 (since we start with 'a').\\- After the recursive calls complete, print each string in `res`.\\
\vspace{-0.2cm}     
\noindent\rule{\textwidth}{1pt}
\caption{\textbf{Reasoner Prompt Example}: Example of the prompt we are using to Fine-tune the \textbf{Reasoner} (GPT-3.5) to generate explanations of the solution. The content from Assistant Prompt is the \textbf{Explainer}'s response to the query in Figure \ref{fig:example_explainer}}
    \label{fig:example_reasoner}
\end{figure*}

\begin{figure*}
\vspace{-1.2cm}
\footnotesize
\noindent\rule{\textwidth}{1pt}
\textbf{System Prompt:} You are an expert in algorithm and programming.\\
\noindent\rule{\textwidth}{0.5pt}
\textbf{User Prompt:} 
Please solve the given algorithmic reasoning problem. You are also given a {Hint} on how to solve the problem. Analyze problem and hint, think carefully before implementing. The ultimate goal is to develop a complete Python 3 program to solve the problem. The program should be designed to read input and produce output in the specified format mentioned in problem. Enclose the code within triple backticks (```) for clarity. Keep in mind that this code is intended for direct submission to an online judge system, so don't include test cases.\\ Here's the Problem:\\
A subarray is a continuous part of array.
Yarik recently found an array a of n elements and became very interested in finding the maximum sum of a non empty subarray. However, Yarik doesn't like consecutive integers with the same parity, so the subarray he chooses must have alternating parities for adjacent elements. \\
For example, [1, 2, 3] is acceptable, but [1, 2, 4] is not, as 2 and 4 are both even and adjacent.
You need to help Yarik by finding the maximum sum of such a subarray.\\
Input\\
The first line contains an integer $t (1 \leq t \leq 10^4)$— number of test cases. Each test case is described as follows. \\
The first line of each test case contains an integer n $(1 \leq n \leq 2 ·10^5)$— length of the array.\\
The second line of each test case contains n integers $a_1, a_2, …, a_n~(-10^3 \leq a_i \leq 10^3)$— elements of the array.\\
It is guaranteed that the sum of n for all test cases does not exceed $2 ·10^5$.\\
Output\\
For each test case, output a single integer— the answer to the problem.\\
\noindent\rule{\textwidth}{0.5pt}
\textbf{Here's the {Hint} on how to solve the problem:} \\
Let's analyze the problem and think how to solve it.
Let's walk through the problem and think about how to solve it first.\\
Conceptual Evolution:\\
To arrive at this solution, one might start by considering the brute force approach of checking all possible subarrays, which would be inefficient. Recognizing that the problem exhibits optimal substructure and overlapping subproblems, dynamic programming becomes a natural choice. The intuition is that the maximum sum of a subarray ending at index `i` can be extended by including `a[i]` if it maintains the alternating parity condition. This insight leads to the iterative update of `dp` and the tracking of the maximum sum.\\
Key to Solution:\\
The "aha!" step is realizing that dynamic programming can be used to efficiently track the maximum sum of subarrays ending at each index, while maintaining the alternating parity condition.\\
Solution Description:\\
The solution uses dynamic programming to keep track of the maximum sum of subarrays ending at each index, while ensuring the alternating parity condition. The key insight is that the maximum sum of a subarray ending at index `i` can be derived from the maximum sum of a subarray ending at index `i-1` if the parity condition is satisfied. The `dp` array is used to store these intermediate results, and the `ans` variable is updated to keep track of the maximum sum found so far.\\
Common Pitfalls:\\
- Not considering the case where the array has only one element, which should be handled separately.\\
- Misunderstanding the alternating parity condition and not updating the `dp` array correctly.\\
- Forgetting to initialize the `dp` array with a value that does not affect the maximum sum (e.g., 0).\\
- Overlooking the possibility of negative numbers in the array, which could lead to incorrect updates of the `dp` array.\\
Step-by-Step Solution Explanation:\\
- Read the number of test cases `t`.\\
- For each test case:\\
  - Read the length of the array `n`.\\
  - Read the elements of the array `a`.\\
  - Initialize a dynamic programming (DP) array `dp` with a length of `n` and fill it with a large negative number (e.g., $-10^9$) to represent the maximum sum of a subarray ending at each index.\\
  - Initialize a variable `ans` to 0, which will hold the maximum sum found.\\
  - If the array has only one element, set `ans` to that element's value.\\
  - Otherwise, iterate over the array starting from the second element:\\
    - If the current element and the previous element have different parities and they are not the first two elements of the array, update the `dp` array at the current index with the maximum of the current element and the sum of the current element and the `dp` value at the previous index.\\
    - If the current element and the previous element have the same parity, update the `dp` array at the current index with the current element (as the previous subarray cannot be extended).\\
    - Update `ans` with the maximum of its current value and the `dp` value at the current index.\\
  - Print the value of `ans` for the current test case.

\noindent\rule{\textwidth}{1pt}
\caption{\textbf{Coder Prompt Example}: Example of the prompt we are using to instruct the \textbf{Coder} generate code given the problem and the verbal solution(also included in this figure). This is the full version for what's in Table \ref{tab:casestudy}}
    \label{fig:example_coder}
\end{figure*}

\end{document}